*Article*

# Tab2vox: CNN-Based Multivariate Multilevel Demand Forecasting Framework by Tabular-To-Voxel Image Conversion

Euna Lee [1,2,*], Myungwoo Nam [1] and Hongchul Lee [1,*]

[1] School of Industrial and Management Engineering, Korea University, Seoul 02841, Korea
[2] Center for Defense Resource Management, Korea Institute for Defense Analyses, Seoul 02455, Korea
\* Correspondence: ealee@korea.ac.kr (E.L.); hclee@korea.ac.kr (H.L.)

**Abstract:** Since demand is influenced by a wide variety of causes, it is necessary to decompose the explanatory variables into different levels, extract their relationships effectively, and reflect them in the forecast. In particular, this contextual information can be very useful in demand forecasting with large demand volatility or intermittent demand patterns. Convolutional neural networks (CNNs) have been successfully used in many fields where important information in data is represented by images. CNNs are powerful because they accept samples as images and use adjacent voxel sets to integrate multi-dimensional important information and learn important features. On the other hand, although the demand-forecasting model has been improved, the input data is still limited in its tabular form and is not suitable for CNN modeling. In this study, we propose a Tab2vox neural architecture search (NAS) model as a method to convert a high-dimensional tabular sample into a well-formed 3D voxel image and use it in a 3D CNN network. For each image representation, the 3D CNN forecasting model proposed from the Tab2vox framework showed superior performance, compared to the existing time series and machine learning techniques using tabular data, and the latest image transformation studies.

## 1. Introduction

### 1.1. Background and Purpose

Understanding the process by which demand is created, identifying the cause of demand generation, and considering these when demand forecasting can help develop demand forecasting technology. In particular, this contextual information can be very useful in demand forecasting for consumer goods with high demand volatility or spare parts with intermittent demand patterns [1].

Most causes of demand are unclear because seasonal factors, regional factors, and user group factors are all integrated. Therefore, it is necessary to determine the explanatory variables by decomposing them into multi-levels, such as year, region, and user group, as shown in Figure 1. That is, for each sample, the following is generated: k (number of explanatory variables) × l (number of clusters in level 1) × m (number of clusters in level 2) × n (number of clusters in level 3) dimensional features to be considered for each sample. More information in the forecasting process requires more parameter estimation, so more complex predictive models and methods are needed [2].

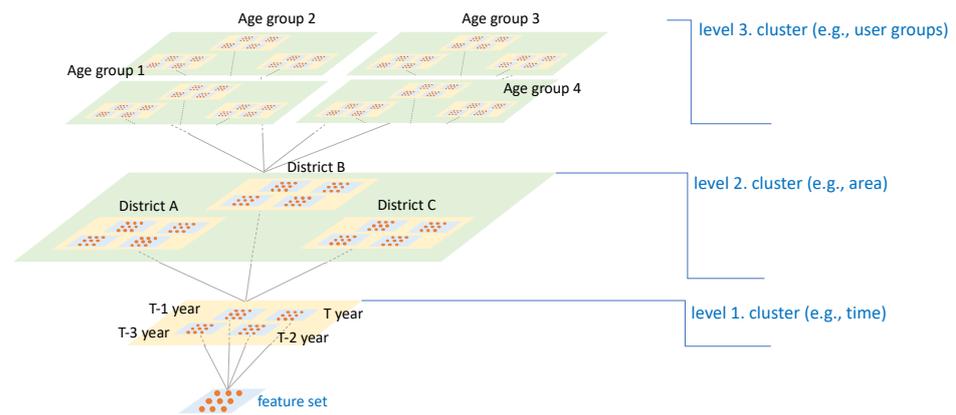

**Figure 1.** Decomposition of demand explanatory multivariate variables by level.

Convolutional neural networks (CNNs) demonstrate high performance in classification and prediction using image data, and they have been successfully used in many applications, such as image and video recognition [3–6], medical image analysis [7,8], natural language processing [9], and speech recognition [10]. However, the tabular data used for demand forecasting is not suitable for CNN modeling, so the application of CNNs in the demand forecasting domain is still limited.

In order to overcome this limitation, it is necessary to convert a tabular feature set into an image, and in this process, it is necessary to develop a method that can effectively perform feature arrangement. Therefore, in this study, we propose a Tab2vox framework that reads spatial context information through 3D CNNs, reflects the multivariate multi-level feature set in demand forecasting, and evaluates its effect (See Figure 2).

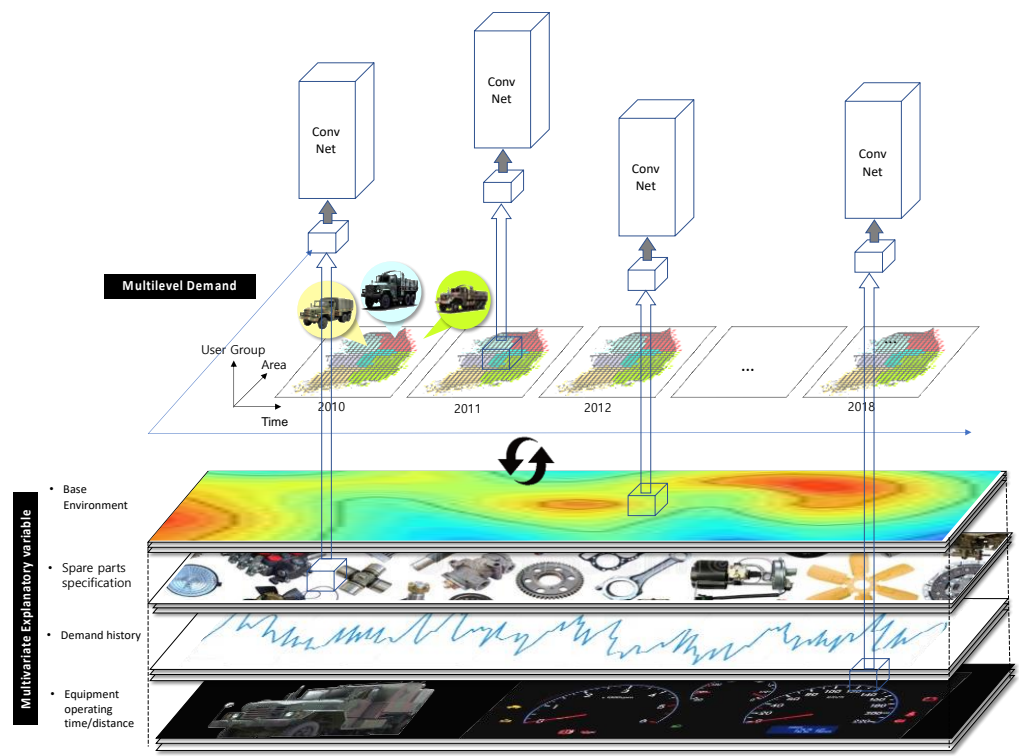

**Figure 2.** An example figure where the 3D CNN spatially reads multivariate multilevel features and extracts a feature map. Demand can be decomposed into levels by year, region, and vehicle model, and the demand explanatory variable is also divided and reflected for each level. The 3D CNN spatially reads the demand cause information across variables between neighboring levels and reflects it in the prediction.

Tab2vox is proposed based on DARTS [11], one of the neural architecture search (NAS) models; DARTS enables the learning of optimized voxel images and 3D CNN structures for demand forecasting based on gradient descent. By extending the DARTS design to solve the classification problem by searching the 2D CNN architecture, we also searched the input-embedding architecture mapping the tabular input data to the optimal voxel image. Furthermore, the deep learning network was modified from the 2D to 3D convolution search space and from the classifier to regressor. These modifications were carried out so that demand forecasting could be performed in consideration of the spatial information of the multivariate multilevel feature.

*1.2. Related Work*

1.2.1. Previous Studies on Demand Forecasting

Demand forecasting research has become much more developed over the past few decades. Traditionally, statistical techniques such as simple exponential smoothing (SES), simple moving average (SMA), ARIMA, and Croston have been mainly used to predict future demand only with past demand, without explanatory variables [12]. Statistical techniques have been widely used for their simplicity and ease of implementation, but they do not reflect demand-influencing factors in forecasting, and their purpose is not to improve forecasting accuracy, but to find and infer the line that best explains the current situation. Therefore, it is difficult to generalize when new patterns of demand appear [13–15].

Data mining is a method to help decision-making analytically by extracting influencers, relationships, patterns, and rules from large-scale data [16]. In the field of demand forecasting, machine learning (ML) approaches such as xGBoost [17], Adaboost [18], and random forests [19] showed excellent performance. Recently, RNN-based [20] algorithms such as LSTM [21] and DeepAR [22], which are specialized for time series data, and techniques using deep learning, such as Transformer-based Tabnet [23] and TabTransformer [24], have been utilized a lot [25].

However, the RNN-based algorithm is effective for data with a sufficiently long time series, and the transformer-based model has many hyperparameters to be tuned with a high level of expertise among deep learning models, so it is not yet widely applied.

Various prediction methodologies were applied by industries, such as the medical [26], resource [27–30], economic [31], and military industries [32], but there have been no studies of CNN deep learning network applications using image data.

1.2.2. Studies on Converting Tabular Data into Images

The case of converting tabular format data into images for CNN deep learning-based prediction has been proposed for gene classification in several recent studies in the biotechnology field, but none in the field of demand forecasting.

The first proposed OmicsMapNet [33] developed a method for converting genetic data into 2D images for tumor grade prediction in cancer patients. However, ontology knowledge extracted from gene and genome encyclopedias was used for feature hierarchical rearrangement, and there is a limitation in that it is difficult to utilize if there is no domain knowledge about the corresponding dataset.

Several studies have proposed methods for projecting n feature vectors into 2D space by reducing them to p(<n) dimensions using dimension reduction methodology [34–36]. When dimensionality reduction is performed, high-dimensional features are displayed in a space that can be visually recognized by humans, so it has the advantage of intuitively helping to understand data and reducing the number of features. However, since the dimension-reduced features have completely different values from the existing features, information loss occurs from the original data. Therefore, it is difficult to interpret the extracted variables [37].

IGTD [38] took a slightly different approach of converting genetic data into a two-dimensional image and predicting an anticancer drug reaction with CNNs. It generates an image by creating an R matrix representing the correlation between features and a Q matrix representing the distance between image coordinates. By exchanging the feature positions of the R matrix so that the R matrix and Q matrix are equal, the highly correlated features are optimized to be located nearby.

1.2.3. Research Gaps

As a result of reviewing the previous studies presented in Sections 1.2.1 and 1.2.2, a research gap was found. It can be summarized as follows:

- Although demand forecasting methodology has become increasingly sophisticated, studies applying the CNN architecture using image data have not yet been conducted.
- Recently, there have been attempts to express tabular data as images in some studies in the field of biotechnology, but image transformation and an evaluation of predictive models were performed independently. That is, it was not a method in which optimal image conversion was performed by linking with the loss function (predictive measurement accuracy).
- In addition, previous studies converting tabular features into images have extracted features using 2D CNNs in the form of integrated demand sources; however, a high-dimensional CNN that integrates and reflects multi-dimensional information between subdivided levels is required.

This study differs from existing demand forecasting studies because it uses image data. It differs from the existing tabular-to-image conversion studies because it applies image transformation directly linked to loss, and uses 3D CNNs to reflect the aspects of multilevel features.

*1.3. Organization of This Paper*

The rest of this paper contains five sections.

Chapter 2 describes Tab2vox, a 3D CNN-based demand forecasting model that converts tabular data into 3D voxel images. Chapter 3 introduces the research framework applied to the military vehicle spare parts case. Chapter 4 compares different methods related to forecast accuracy and examines the usefulness of using 3D image data and 3D CNN models for demand forecasting. Finally, Chapters 5 and 6 conclude the paper by summarizing the study and explaining the limitations of this paper to be explored in future research.

**2. Materials and Methods**

*2.1. Neural Architecture Search (NAS)*

Automated machine learning (AutoML) is a research field that automates the entire machine learning pipeline, from data preparation to model evaluation. Recently, neural architecture search (NAS), which automatically finds network architectures such as layer type and number of layers, has attracted attention [39]. NAS uses the objective function to minimize validation loss or forecast accuracy, and it is trained by defining which architecture to use as a search space.

There are four representative approaches to NAS: reinforcement learning, evolutionary, Bayesian, and gradient-based algorithms, and which algorithm to choose remains a manual problem. All other NAS methodologies before DARTS [11] performed architecture searches in such a way that the model was evaluated after sampling the architecture candidates from the discrete search space, and then the procedure of sampling and evaluating other architecture candidates was repeated. In this process, evaluating thousands or tens of thousands of structures results in a huge search cost. To overcome this problem, DARTS applies continuous relaxation to the discrete search space to make it differentiable and to perform a fast and efficient architecture search through gradient descent-based optimization.

In many cases, model run time has a particular importance in demand forecasting, such as in the case of consumer goods that rely on checking the sale price, weather, season, and event status in real time, or the defense domain that requires combat-readiness. In this study, based on the DARTS algorithm, which has the fastest run time among NAS methodologies, we propose a demand forecasting framework that (1) searches for an optimal method for converting tabular data into 3D voxel images, and (2) searches for an optimal 3D CNN network.

*2.2. Proposed Method: Tab2vox*

Tab2vox extends the existing DARTS, that ordinarily searches only CNN network architectures, to search architectures that map tabular input data to images. This section explains how Tab2vox performs NAS. There are two architectures that Tab2vox searches: an input representation architecture that maps tabular format data to a 3D voxel image, and a 3D CNN architecture.

The first stage searches the input-embedding architecture, which includes a four-step workflow.

(i) As an example, suppose that we do not know in what order to map the features of each level to the 3D voxel image, as shown in Table 1(a). First, the image is initialized by arranging the feature set for each level in the tabular form on each arbitrary axis of the voxel image. In tabular format data, the order in which the feature set is listed is not important, but in the voxel image, adjacent cells are meaningful.

(ii) Next, clustering is performed on each feature set for each level, as shown in Table 1(b). The point of this study is to determine how to map features divided into k explanatory variables (level 0), l region (level 1), and m users (level 2) to each axis of the 3D voxel image. Next, the number of cases k! × l! × m! should be defined as the search space and searched. Therefore, in this study, first, features by level were grouped to reduce the search space. In this case, grouping may be performed using k-means [40], knn [41], hierarchical clustering [42], etc., which are commonly used cluster analysis techniques.

Assume that, after performing clustering, groups for each level are classified into, for example, 5, 3, and 2 clusters, respectively. Even with clustering, the number of voxel images that can be created is 5! × 3! × 2! = 1440, and if you train the model weight parameters for all these architectures and repeat this process, a large search cost will occur.

(iii) Therefore, in Tab2vox, as shown in Table 1(c), the output value of the input-embedding network is defined as a weighted sum by candidate mapping images to make it differentiable. This applies the idea of DARTS. For the operation o that connects tabular data to image i, $\alpha_o^i$ is considered a weight, multiplied by o(x) by the softmax weight, and all the values are then added to calculate the weight of candidate image operations. That is, the larger $\alpha_o^i$ is, the more the operation is affected, and the closer that $\alpha_o^i$ is to 0, the more the operation does not play any role. The weights defined in this way become continuous variables, meaning that these weights can be trained to minimize validation loss using gradient descent.

Secondly, CNN architecture is also searched by using the network the architecture search concept of DARTS as it is. For the purpose of this study, only the operation set was extended to three dimensions by referring to existing studies to read multi-dimensional information between levels [43–45]. DARTS also constructs a CNN model by creating cells like VGG [46] and RestNet [47] and stacking cells repeatedly. At this time, the number of cells, the number of nodes in the cell, and the normal/reduction cell configuration method are all based on the assumptions of DARTS.

Finally, for the regressor, the network at the end is flattened, connected, and passed to the fully connected network for prediction. DARTS was originally designed to solve the classification problem, but in this study, the output of the last linear layer was converted into one node so that the regression task could be performed.

By training the Tab2vox network designed in this way, the optimal unique image in the form of a 3D voxel for each sample and the optimal 3D CNN network are obtained. Table 1 summarizes the differences between Tab2vox and the existing DARTS. The Supplementary Materials includes detailed information on the Tab2vox framework.

**Table 1.** DARTS vs Tab2vox.

| | DARTS | Tab2vox |
|---|---|---|
| Input Embedding Architecture Search | none | Tabular to image input embedding search space = { (explanatory feature cluster order, area cluster order, user cluster order) } |
| CNN Architecture Search | 2D CNN search space = { none, avg_pool_3x3, max_pool_3x3, skip_connect, sep_conv_3x3, sep_conv_5x5, sep_conv_7x7, dil_conv_5x5, conv_7x1_1x7 } | 3D CNN operations search space = { none, avg_pool_3x3x3, max_pool_3x3x3, skip_connect, conv_1x1x1, conv_3x3x3, conv_5x5x5, sep_conv_3x3x3, dil_conv_3x3x3, conv_5x5x5, conv_1x3x3, conv_3x1x1 } |
| Task | Classification | Regression |

## 3. Case Study of MND (Ministry of National Defense, Republic of Korea) Spare Parts Dataset by Tab2vox

### 3.1. Datasets

To verify the performance of Tab2vox, we proposed a research framework, as shown in Figure 3. The following sections describe the details of the proposed research framework.

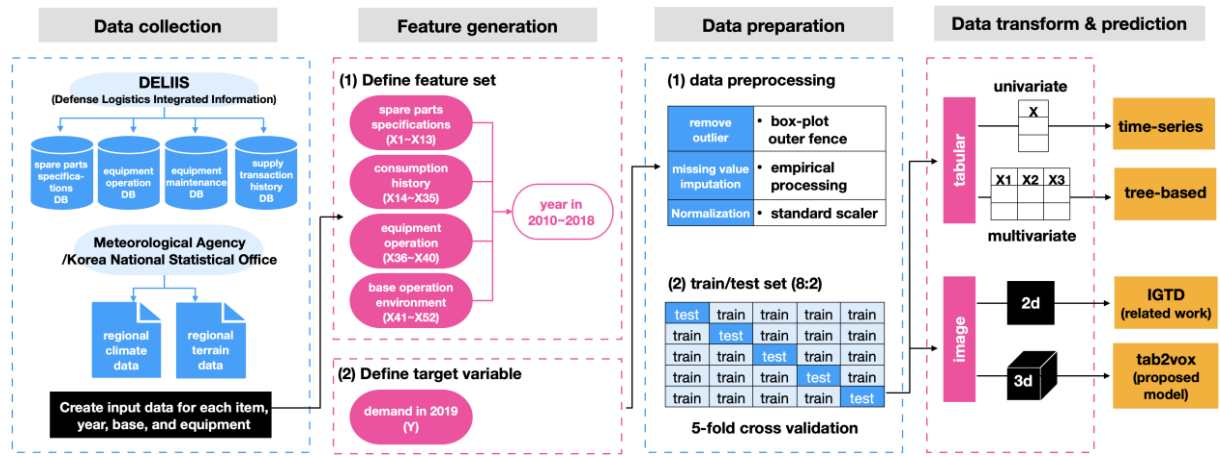

**Figure 3.** The research framework proposed by this paper.

#### 3.1.1. Data Collection

For the analysis, the ROK K-511 vehicle spare parts dataset was used. Military vehicles have the largest number of weapon systems in operation in the ROK and have a diverse operating area, so relatively sufficient samples and explanatory variables can be used. In addition, the analysis was conducted on 2290 types of spare parts that were in demand between 2010 and 2019.

The K-511 Military Vehicle dataset used in the analysis was collected from the Defense Logistics Integrated Information System (DELIIS). This system stores information related to the ROK military's weapon systems and spare parts, such as supply transaction history, maintenance history, and equipment operation record. All data available since 2010 when the system was first developed were used in this study.

Furthermore, topographical and climate data were collected from the National Statistics Portal of the Korea National Statistical Office (https://kosis.kr/ accessed on 1 January 2022) and the Data Open Portal of the Korea Meteorological Administration (https://data.kma.go.kr/ accessed on 1 January 2022), respectively, and they were used as explanatory features for units in operation K-511.

#### 3.1.2. Feature Generation

The collected data, such as the daily spare parts replacement history, supply/maintenance history, etc., were directly processed and used for analysis by year, unit, and equipment level. As mentioned at the beginning of this paper, the cause of most demand is unclear because seasonal factors, regional factors, and user group factors are all integrated. It is necessary to consider decomposing features into multi-level decomposition.

In addition to the basic model used to transport military supplies and troops, K-511 vehicles are operated as various derivatives for maintenance, oil transport, command and control, water supply, decryption, and corrosion transport. Vehicles operating with the same mission have a high failure probability of similar spare parts, and conversely, vehicles with different characteristics and operating concepts will show different demand patterns, even for the same item.

Even with the same spare part, the spare parts of vehicles operating in coastal areas can be more corroded due to sea winds, and the spare parts of equipment operating in mountainous areas—such as track spare parts—can be frequently in demand due to the rugged terrain.

In addition, yearly data can reflect time-series characteristics of spare parts failure.

Table 2 lists and explains 52 variables created for military vehicle demand forecasting. Variables are divided into 13 spare part specification variables, 23 demand history variables, and 16 equipment operating environment variables. The explanatory feature set includes all variables used in previous studies. In previous studies, features were divided up to the year level and the base level at the maximum, and in this study, features were extended to the equipment level and used for prediction.

The explanatory variable consists of data from 2010–2018, and demand in 2019 was defined as the target variable.

**Table 2.** Definition of explanatory features used for forecasting the spare parts demand of the military vehicle system in year t (t = 2010–2018).

| Category | Feature | Definition | Type | Range | Level [1] 1 | Level [1] 2 | Level [1] 3 |
|---|---|---|---|---|---|---|---|
| Spare parts specifications | X1 | Unit price of spare parts (KRW) | numerical | [3, 13,519,541] | | | |
| | X2 | Length of spare parts (cm) | numerical | [0, 350] | | | |
| | X3 | Width of spare parts (cm) | numerical | [0, 520] | | | |
| | X4 | Height of spare parts (cm) | numerical | [0, 138] | | | |
| | X5 | Weight of spare parts (kg) | numerical | [0, 1800] | | | |
| | X6 | Volume of spare parts (m$^3$) | numerical | [0, 3] | | | |
| | X7 | Quantity Per Application (QPA): Number of spare parts installed per weapon system | numerical | [1, 168] | | ✓ | |
| | X8 | Base Repair Percent (BRP): Average percentage of spare parts repaired at bases out of maintenance requests | numerical | [0, 1] | | | |
| | X9 | Depot Repair Percent (DRP): Average percentage of spare parts repaired at depot out of maintenance requests from bases | numerical | [0, 1] | | | |
| | X10 | Condemnation Percentage (ConPct): Average percentage of spare parts discarded as non-repairable by depot | numerical | [0, 1] | | | |
| | X11 | Average repair cost of spare parts at depot base level | numerical | [2, 13,699,405] | | ✓ | |
| | X12 | Average repair cost of spare parts at depot maintenance level | numerical | [0.9, 7,229,964] | | ✓ | |
| | X13 | Procurement Lead Time (PROLT): Lead time, which is the period time (in days) from the purchase request to the receipt completion for spare parts | numerical | [18, 999] | | | |
| Consumption history | X14 | Monthly spare parts demand average in year t | numerical | [0, 243] | ✓ | ✓ | ✓ |
| | X15 | Monthly spare parts demand median in year t | numerical | [0, 115] | ✓ | ✓ | ✓ |
| | X16 | Monthly spare parts demand deviation in year t | numerical | [0, 182,992] | ✓ | ✓ | ✓ |
| | X17 | Daily spare parts demand average in year t | numerical | [0, 8] | ✓ | ✓ | ✓ |
| | X18 | Daily spare parts demand median in year t | numerical | [0, 1] | ✓ | ✓ | ✓ |
| | X19 | Daily spare parts demand deviation in year t | numerical | [0, 4749] | ✓ | ✓ | ✓ |
| | X20 | The amount of spare parts consumption in January in year t | numerical | [0, 1] | ✓ | ✓ | ✓ |
| | X21 | The amount of spare parts consumption in February from 2010 to year t | numerical | [0, 1] | ✓ | ✓ | ✓ |
| | X22 | The amount of spare parts consumption in March from 2010 to year t | numerical | [0, 1] | ✓ | ✓ | ✓ |
| | X23 | The amount of spare parts consumption in April from 2010 to year t | numerical | [0, 1] | ✓ | ✓ | ✓ |
| | X24 | The amount of spare parts consumption in May from 2010 to year t | numerical | [0, 1] | ✓ | ✓ | ✓ |
| | X25 | The amount of spare parts consumption in June from 2010 to year t | numerical | [0, 1] | ✓ | ✓ | ✓ |
| | X26 | The amount of spare parts consumption in July from 2010 to year t | numerical | [0, 1] | ✓ | ✓ | ✓ |
| | X27 | The amount of spare parts consumption in August from 2010 to year t | numerical | [0, 1] | ✓ | ✓ | ✓ |
| | X28 | The amount of spare parts consumption in September from 2010 to year t | numerical | [0, 1] | ✓ | ✓ | ✓ |
| | X29 | The amount of spare parts consumption in October from 2010 to year t | numerical | [0, 1] | ✓ | ✓ | ✓ |
| | X30 | The amount of spare parts consumption in November from 2010 to year t | numerical | [0, 1] | ✓ | ✓ | ✓ |
| | X31 | The amount of spare parts consumption in December from 2010 to year t | numerical | [0, 1] | ✓ | ✓ | ✓ |
| | X32 | Average monthly spare parts inventory quantity in year t | numerical | [0, 9150] | ✓ | | ✓ |
| | X33 | Spare parts consumption in year t | numerical | [0, 2919] | ✓ | ✓ | ✓ |
| | X34 | Squared coefficient of variation ($CV^2$): Standard deviation of the demand divided by the average demand for non-zero demand periods. | numerical | [0, 63] | ✓ | | ✓ |
| | X35 | Average inter-demand interval (ADI): Average interval time between two demand occurrences. | numerical | [0, 9] | ✓ | | ✓ |
| | X36 | Mean time between failures (MTBF): | numerical | [1, 9] | ✓ | | ✓ |
| | X37 | Operating distance (in hours) of the weapon system, related to a spare part, in year t | numerical | [0, 3,229,201] | | | ✓ |
| | X38 | Operating time (in hours) of the weapon system, related to a spare part, in year t | numerical | [0, 7,854,334] | | | ✓ |
| | X39 | Oil consumption in year t | numerical | [0, 406,682] | | | ✓ |
| | X40 | Average equipment age in year t | numerical | [2.85, 35] | | ✓ | ✓ |
| | X41 | Number of weapon systems operated in year t | numerical | [0, 3584] | ✓ | ✓ | ✓ |
| | X42 | Average temperature up to year t in the weapon system operating bases | numerical | [21.2, 25] | ✓ | | |
| | X43 | Maximum temperature up to year t in the weapon system operating bases | numerical | [25.3, 30] | ✓ | | |
| Operating environment | X44 | Minimum temperature up to year t in the weapon system operating bases | numerical | [15.9, 22] | ✓ | | |
| | X45 | Average amount of precipitation up to year t in the weapon system operating bases | numerical | [1.2, 17] | ✓ | | |
| | X46 | Average humidity up to year t in the weapon system operating bases | numerical | [61.4, 86] | ✓ | | |
| | X47 | Weapon system utilization rate in year t | numerical | [0.98, 1] | ✓ | | |
| | X48 | Forest ratio to total area (%) | numerical | [23.76, 97] | ✓ | | |
| | X49 | Road pavement ratio to total area (%) | numerical | [3.72, 100] | ✓ | | |
| | X50 | Land ratio to total area (%) | numerical | [0.79, 1] | ✓ | | |
| | X51 | Sea ratio to total area (%) | numerical | [0, 0] | ✓ | | |
| | X52 | Urban ratio to total area (%) | numerical | [0.53, 100] | ✓ | | |

[1] level 1: bases, level 2: equipment, level 3: years.

3.1.3. Data Preparation

For the explanatory features divided by the level created in this way, outliers were removed by applying the outer fence of the box-plot, and the missing value was corrected using similar items and similar base information. In addition, except for the time series model that only uses annual demand data, the data distribution of the existing variables was converted into a normal distribution with a mean of 0 and a variance of 1, and was used as input data after normalization.

For fair comparison, all predictive models were trained and tested with 5-fold cross validation using the same data partition (i.e., training, validation, and testing sets). For the collected data, 60% of the item data was used for model training, 20% was used for validation to avoid overfitting, and the remaining 20% was used to test the prediction performance.

### 3.1.4. Data Representation & Prediction

As shown in Figure 4, the collected data were initially refined into a matrix consisting of m = 1,008,549 rows (= item × unit × equipment × year) and n = 56 columns (=4 level distinguisher + 52 explanatory features). Afterwards, the matrix was used to predict the transformation into univariate and multivariate tabular data and 2D and 3D images according to the model.

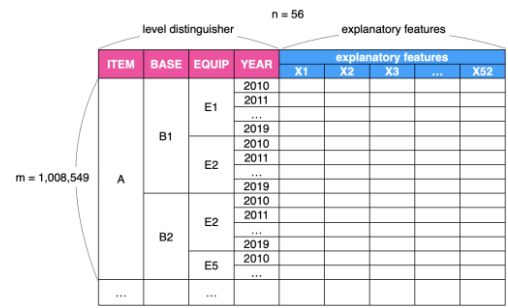

**Figure 4.** Initial ROK military vehicle spare parts demand dataset in tabular format. Demand by item is divided into base level, equipment level, and year level, and then explanatory features are considered.

### 3.2. Evaluation with Comparisons

#### 3.2.1. Overview

The main goal of this study is to show that CNN architecture can be applied using non-image data through Tab2vox implementation. To prove the effectiveness of the Tab2vox demand forecasting framework, the performance was evaluated in comparison with the time series, regression, and ensemble methods, which have typically been adopted as benchmarking methods in previous studies. For the hyperparameters of comparative models, random grid search [48] optimization was performed to ensure a fair comparison with the best performance for each model. Next, 5-fold cross validation was performed to compare the average prediction accuracy for five test sets.

In addition, the performance was compared with IGTD, which is achieving state-of-the-art (SOTA) results among existing studies converting tabular data into image form.

#### 3.2.2. Tab2vox Architecture for the MND Dataset

By applying Tab2vox NAS to the MND dataset, the optimal image-embedding architecture and CNN architecture—as shown in Figure 5—were derived and used for comparison with other prediction techniques.

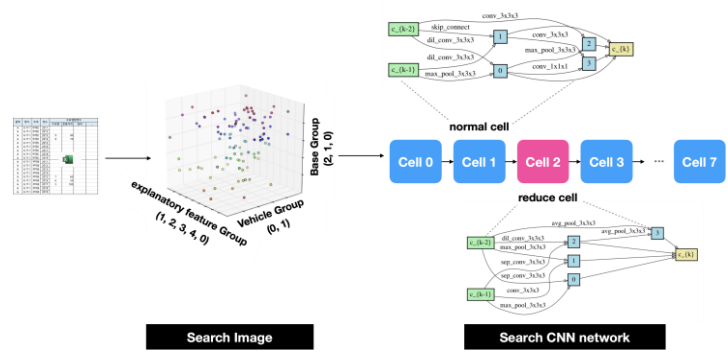

**Figure 5.** Architecture of the tabular-to-image conversion and the convolutional neural network (CNN) used for predicting the spare parts demand for military vehicles based on Tab2vox.

3.2.3. Machine Learning Methods

Table 3 shows the prediction models analyzed together for this study's Tab2vox performance evaluation; all models use tabular input data.

To implement each technique, the machine learning technique used the PyCaret AutoML module and IGTD used the code from GitHub.

- Machine Learning: the PyCaret Python package was used with the tuned parameters.
- IGTD: the public IGTD source code was used, which is available at https://github.com/zhuyitan/IGTD (accessed on 1 January 2022).

**Table 3.** Machine learning methods used with their abbreviations.

| Time Series | Regression | Ensemble |
|---|---|---|
| Arithmetic Mean (am) | Linear Regression (lr) | Random Forest (rf) |
| Simple Exponential Smoothing (ses) | Decision Tree (dt) | AdaBoost (ada) |
| Weighted Moving Average (wma) | Ridge Regression (ridge) | Gradient Boosting Regressor (gbr) |
| | Bayesian Ridge (br) | Extreme Gradient Boosting (xgboost) |
| | Lasso Regression (lasso) | Light Gradient Boosting Machine (lightgbm) |
| | ElasticNet (en) | CatBoost Regressor (catboost) |

## 4. Results

### 4.1. Evaluation Metrics

To evaluate the demand forecast performance between the predicted value $F_i$ and the actual value $A_i$ for item i derived by the models, three measures commonly used to evaluate the forecast model—the root mean square error (RMSE), the mean absolute error (MAE), and the min–max accuracy method—were used. The calculation formula for each measure is as follows (1)–(3).

The RMSE is calculated by rooting the mean of the squares of the difference between the predicted and actual values. The RMSE measure can be very useful for the dataset used in this study because it reduces the scale when the error has a very large value. The MAE is measured by averaging the absolute value of the difference between the predicted value and the actual value, and it is effective when used in an evenly distributed error. For both the RMSE and the MAE, the smaller the value, the smaller the prediction error, which means a better performance model.

The min–max accuracy method measures how close the predicted value is to the actual value by taking the average of the minimum and maximum values between the actual and predicted values. The min–max accuracy ranges between 0 and 1, and if the min–max accuracy is 1, the actual and predicted values are equal. Unlike RMSE and MAE, min-max accuracy has the advantage of easy interpretation because it measures the ratio of the observed value to the predicted value regardless of the scale.

Formula of the RMSE:

$$\text{RMSE} = \sqrt{\frac{\sum_{i=1}^{n}(F_i - A_i)^2}{n}} \quad (1)$$

Formula of the MAE:

$$\text{MAE} = \frac{1}{n}\sum_{i=1}^{n}|F_i - A_i| \quad (2)$$

Formula of the min–max accuracy:

$$Min - Max\ Accuracy = mean\left(\frac{\min(A_i, F_i)}{\max(A_i, F_i)}\right) \quad (3)$$

In the above equation, $i$ denotes each sample, $F_i$ denotes a predicted value by each model, and $A_i$ denotes an actual value.

*4.2. Prediction Accuracy by a Single Model*

Table 4 and Figure 6 show the demand forecasting accuracy derived using various data representations and forecasting models. To comprehensively evaluate the proposed Tab2vox model, we compared the mean of performance metrics for five test sets using time series and ML techniques.

Among all methods, Tab2vox showed the best performance, showing 63.1% in terms of the average min–max prediction accuracy of all items. Next, decision tree, a regression method, showed good performance with 62.9%; random forest, an ensemble method, with 54.5%; XGBoost with 50.6%; and Catboost with 49.3%. For reference, since RMSE and MAE have scale-dependent error characteristics, they are presented as auxiliary indicators of min-max prediction accuracy, meaning that Tab2vox showed better performance in items with relatively small average demand size.

On the other hand, other time series techniques, regression models, and classical ensemble techniques did not show good performance in predicting the demand for spare parts. IGTD, a state-of-the-art model applying CNN architecture by converting tabular-format data into images, also did not perform well. It is thought that this may be due to the fundamental difference in data characteristics between the spare parts dataset of the intermittent demand pattern and the dense genetic dataset. Further analysis is required.

**Table 4.** Comparison of the military vehicle demand prediction performance of different data representations and prediction models. The number is the average and standard deviation of accuracy across five cross-validation trials. Bold indicates the best model with the highest min–max accuracy.

| Data Representation | Prediction Model | | Min-Max | | RMSE | | MAE | |
|---|---|---|---|---|---|---|---|---|
| | | | Mean | Std | Mean | Std | Mean | Std |
| Tabular data | Time Series | Arithmetic Mean | 0.343 | 0.015 | 486.92 | 84.67 | 110.98 | 10.05 |
| | | Simple Exponential Smoothing | 0.486 | 0.010 | 249.15 | 63.51 | 56.47 | 9.36 |
| | | Weighted Moving Average | 0.449 | 0.012 | 389.38 | 77.37 | 83.57 | 9.39 |
| | Regression | LinearRegression | 0.333 | 0.043 | 140.57 | 24,799.10 | 41.16 | 1371.85 |
| | | DecisionTree | 0.629 | 0.024 | 166.41 | 40.27 | 36.40 | 5.97 |
| | | Ridge | 0.332 | 0.044 | 140.62 | 52.31 | 41.25 | 6.76 |
| | | BayesianRidge | 0.334 | 0.044 | 140.62 | 26.69 | 41.09 | 4.70 |
| | | Lasso | 0.230 | 0.022 | 179.27 | 28.70 | 47.59 | 4.37 |
| | | ElasticNet | 0.295 | 0.024 | 213.86 | 39.06 | 52.70 | 6.17 |
| | Ensemble | RandomForest | 0.545 | 0.021 | 148.61 | 37.62 | 34.58 | 6.09 |
| | | AdaBoost | 0.094 | 0.031 | 305.53 | 1622.54 | 182.08 | 898.37 |
| | | GradientBoosting | 0.388 | 0.025 | 154.79 | 29.27 | 34.17 | 4.10 |
| | | XGBoost | 0.506 | 0.016 | 133.17 | 33.40 | 31.42 | 5.36 |
| | | LGBM | 0.468 | 0.019 | 150.36 | 30.80 | 32.42 | 4.89 |
| | | CatBoost | 0.493 | 0.006 | 156.24 | 37.63 | 33.51 | 5.47 |
| Images | IGTD (2D images) | 2D-CNN | 0.171 | 0.006 | 485.45 | 157.85 | 112.58 | 28.71 |
| | Tab2Vox (3D images) | 3D-CNN | 0.631 | 0.014 | 518.81 | 160.65 | 92.28 | 22.86 |

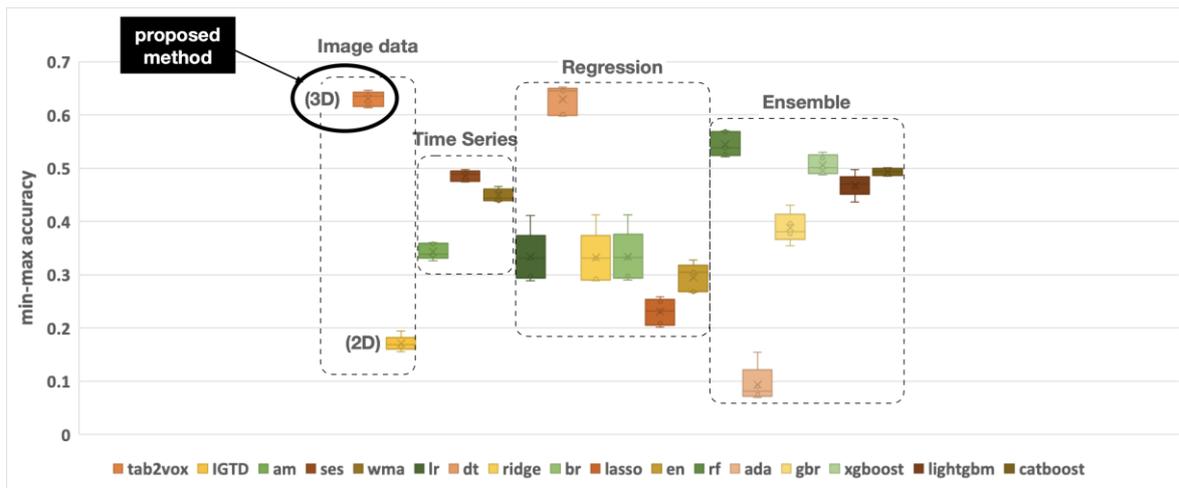

**Figure 6.** A boxplot showing the accuracy deviation of five test sets for each model.

*4.3. Ensemble Approach*

4.3.1. Model Selection Problem

"All models are wrong." This is a quote from the British statistician George EP Box [49,50]. This means that there is no single model that can completely explain all cases. In this chapter, we tried to determine whether better performance could be secured with the most effective model combination for each item group, rather than applying a single model to all items, as shown in Figure 7.

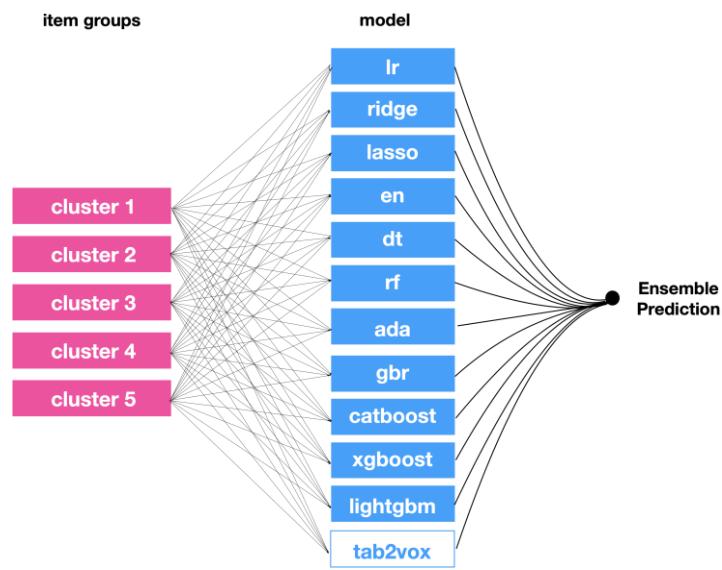

**Figure 7.** An ensemble approach to explore the combination of predictive models by item group.

In this analysis, 458 randomly selected items were analyzed. The time series method that does not consider demand explanatory features, and IGTD—which had the lowest prediction accuracy—was excluded. Only the Tab2vox model and multivariate machine learning method were included. The forecast values for each item of the individual model were created as a new dataset, and the average forecasts of the item group were used as metadata.

First, cluster analysis was performed to classify item groups. Table 5 shows the results of cluster analysis. Four item groups were classified through hierarchical clustering based on the item unit price and average annual demand, the features most correlated with demand generation. The number of clusters was set in consideration of the silhouette score, a metric that can quantitatively evaluate clustering.

**Table 5.** Clustering result on four item groups and their criteria.

| Item Group | Clustering Rules | | Number of Items |
| --- | --- | --- | --- |
| | Item Cost (IC) | Annual Demand (AD) | |
| Item group 1 | <8M KRW | <1.3 | 433 |
| Item group 2 | | ≥1.3 | 19 |
| Item group 3 | | ≥4 | 4 |
| Item group 4 | ≥8M KRW | - | 2 |

Next, an integer programming (IP) model—which searches for the optimal combination of predictive models for each item group—was proposed based on the assumptions and notation in Table 6. The IP model was coded in the General Algebraic Modeling System (GAMS) 39.2.1 with CPLEX 22.1.0 on a Core i7 processor and 16GB RAM.

**Table 6.** Notation for model selection.

| Symbol | Description |
| --- | --- |
| Sets | |
| | Set of item groups = {0, 1, …, n} |
| | Set of prediction models = {0, 1, …, 12} |
| Parameters | |
| $c_i$ | Number of items in item group $i$, $i \in I$ |
| $acc_{ij}$ | Prediction accuracy of model j in item group $i$, $i \in I$, $j \in J$ |
| $std_{ij}$ | Standard deviation between the prediction accuracies of 5-fold test sets in model j in item group $i$, $i \in I$, $j \in J$ |
| $t_j$ | Run time of model $j$, $j \in J$ |
| $c$ | Total number of test items |
| $T$ | Total run time |
| Decision variables | |
| $x_{ij}$ | Binary decision variable, equal to 1 if the item group $i \in I$ selects the prediction model $j \in J$. |

$$Maximize \sum_{i \in I} \sum_{j \in J} (c_i / c) \, acc_{ij} \, x_{ij} \quad (4)$$

$$\sum_j x_{ij} = 1, \quad i \in I \quad (5)$$

$$\sum_{i \in I} \sum_{j \in J} (c_i/c) \, t_j \, x_{ij} \leq T \quad (6)$$

$$x_{ij} \in \{0,1\} \quad for \quad \forall_{i,j} \quad (7)$$

Equation (4) searches for a combination of predictive models that maximize the accuracy of each item group as an objective function. In constraint (5), one item group selects one predictive model, and constraint (6) sets the upper limit of the model run time of all item groups. Finally, Equation (7) is a constraint for setting a decision variable.

As a result of exploring the optimal predictive model combination for the four item groups, it was found that 65.58% accuracy could be achieved if the proposed Tab2vox model was used with dt, lasso, and xgboost, as shown in Table 7. This result shows an improvement of 2.58%, compared to when Tab2vox was applied as a single technique to all items.

Table 7. Min–max accuracy for optimal item group and optimal model combination.

| Cluster | Number of Items | ada | br | catboost | dt | en | gbr | Lasso | lightgbm | lr | rf | Ridge | xgboost | Tab2vox |
|---|---|---|---|---|---|---|---|---|---|---|---|---|---|---|
| 1 | 433 | 0.074 | 0.322 | 0.488 | 0.630 | 0.283 | 0.377 | 0.212 | 0.460 | 0.322 | 0.541 | 0.319 | 0.498 | **0.655** |
| 2 | 19 | 0.575 | 0.631 | 0.641 | 0.606 | 0.564 | 0.628 | 0.593 | 0.638 | 0.631 | 0.634 | 0.630 | **0.659** | 0.189 |
| 3 | 4 | 0.028 | 0.275 | 0.366 | **0.767** | 0.218 | 0.485 | 0.385 | 0.556 | 0.222 | 0.644 | 0.259 | 0.671 | 0.435 |
| 4 | 2 | 0.037 | 0.271 | 0.482 | 0.386 | 0.299 | 0.336 | **0.497** | 0.415 | 0.225 | 0.333 | 0.425 | 0.422 | 0.041 |

Total average min–max accuracy: 65.58%.

This time, we analyzed the feasible region of Tab2vox according to the number of item groups, run time constraints, and robustness scenarios.

4.3.2. Optimal Model Combination Considering the Number of Item Group Clusters

In the previous section, the optimal cluster size of the test item group was analyzed as four. However, when using the model in an actual industrial field, it is also possible to use a model for a smaller or larger item group according to the capacity. Figure 8 and Table 8 show the results of deriving the optimal model combination and prediction accuracy according to item group size using the previously defined IP model. The item group classification result and its criteria can be found in the Supplementary Materials.

When the Tab2vox model was used together with the dt, xgboost, lasso, and br models, it was found that a prediction accuracy of 66.85% could be achieved when there were 10 item groups. Just because an item group is currently supplemented by other machine learning models does not mean that Tab2vox cannot work well with that item group. This is because the Tab2vox model applied in this study has an image embedding- and CNN network architecture that has been searched for good performance on average for all items. If the optimal architecture is searched by applying Tab2vox to each item group, Tab2vox can be applied to more item groups.

On the other hand, as the item groups are subdivided and the number of models used increases, management requirements increase accordingly, and there is a possibility that the model combination may be judged as overfitting. Therefore, it is necessary to apply an appropriate item group x model combination, which is beyond the scope of this study.

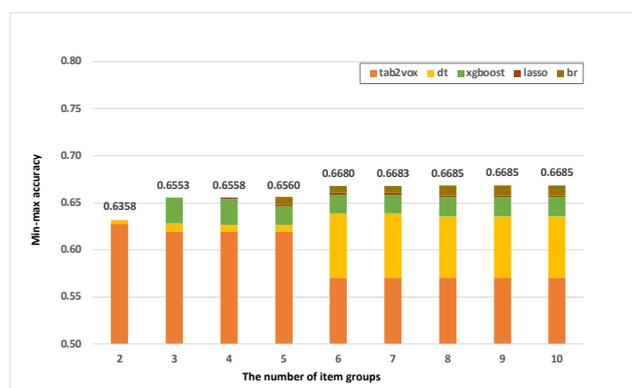

**Figure 8.** Optimal prediction model combination and prediction accuracy graph considering the number of item groups.

**Table 8.** Optimal prediction model combination and prediction accuracy table considering the number of item groups.

| Model Combinations | Number of Item Groups (A) | Number of Models (B) | Workload (A × B) | Min–Max Accuracy |
| --- | --- | --- | --- | --- |
| tab2vox + dt | 2 | 2 | 4 | 0.6358 |
| tab2vox + dt + xgboost | 3 | 3 | 9 | 0.6553 |
| tab2vox + dt + xgboost + lasso | **4 (optimal)** | 4 | 16 | **0.6558** |
| tab2vox + dt + xgboost + lasso + br | 5 | 5 | 25 | 0.6560 |
| tab2vox + dt + xgboost + lasso + br | 6 | 5 | 30 | 0.6680 |
| tab2vox + dt + xgboost + lasso + br | 7 | 5 | 35 | 0.6683 |
| tab2vox + dt + xgboost + lasso + br | 8 | 5 | 40 | 0.6685 |
| tab2vox + dt + xgboost + lasso + br | 9 | 5 | 45 | 0.6685 |
| tab2vox + dt + xgboost + lasso + br | 10 | 5 | 50 | 0.6685 |

4.3.3. Optimal Model Combination According to Run Time Constraints

We looked for the optimal model combination according to the run-time constraint and the feasible region that can use Tab2vox. As shown in Figure 9, if the run-time is given for more than four hours based on the 1832 items of the train dataset used in this analysis, Tab2vox can be useful as it has a high performance of 65.58%.

In this study, due to the difficulty of measuring quantitative indicators, only the pure training time of the model was considered as the run time. However, considering the time spent on feature engineering or hyperparameter tuning in machine learning techniques, Tab2vox can be much more useful.

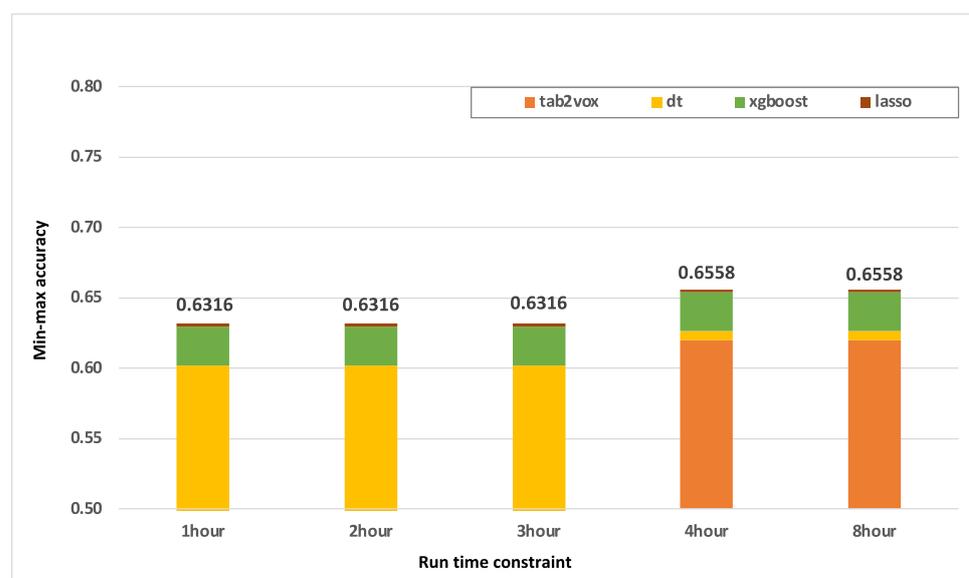

**Figure 9.** Optimal prediction model combination and prediction accuracy considering run time.

### 4.3.4. Optimal Model Combination Considering Robustness

By changing the objective function of the IP model proposed in Section 4.3.2., we examined the optimal model combination considering the robustness. The model robustness was defined as the standard deviation of five test sets set to 5-fold, and the analysis was conducted based on the optimal number of clusters: four item groups. As shown in Figure 10, even considering the ratio of robustness up to half, Tab2vox was found to work effectively in most item groups.

$$Maximize \sum_{i \in I} \sum_{j \in J} W1 \times (c_i/c) \times acc_{ij} \, x_{ij} - W2 \times (c_i/c) \times std_{ij} \, x_{ij} \tag{8}$$

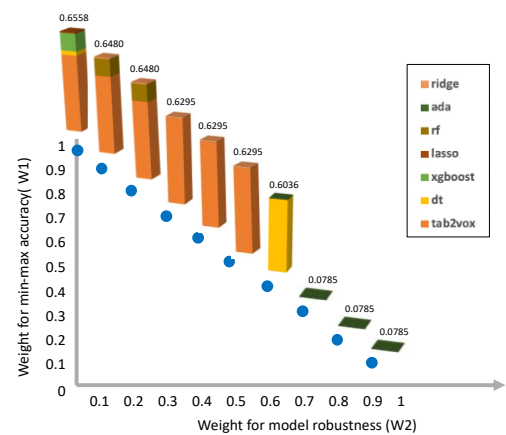

**Figure 10.** Optimal prediction model combination and prediction accuracy considering robustness.

## 5. Discussion

A CNN takes a sample as an image and uses a set of adjacent pixels to learn various features. However, the application of CNNs in the field of demand forecasting has been limited. This is because tabular data, which has been traditionally used for demand forecasting, assumes that the relationship among features is independent. To apply high-dimensional tabular data to CNNs, it is necessary to link non-image information to image-based learning. In this work, we propose an optimal image conversion search framework for using high-dimensional multilevel features as inputs to CNN architecture.

As expected, the mapping of tabular data found by the Tab2vox NAS model to the voxel image space appears to be useful for feature extraction using a 3D CNN architecture. Compared to traditional time series and machine learning, it showed better performance, and it even surpassed the performance of the latest image conversion model proposed in the field of biotechnology. As a result of applying it to the ROK military vehicle spare parts dataset, Tab2vox showed 63.10% performance when applied with a single model to all items; this is the best performance among other baseline models. Furthermore, when Tab2vox was used together with other machine learning techniques, the prediction accuracy in the optimal item group x model combination was 65.58%.

## 6. Conclusions

### 6.1. Implications

The significance of the proposed Tab2vox methodology can be summarized as follows.

First, the proposed demand forecasting framework does not solve all demand forecasting problems. However, it does offer enormous potential because it provides a cornerstone for applying the demand forecasting problem to CNNs, an advanced model with many advantages. Converting similar features into images by inserting them into neighbors is more meaningful and powerful than ignoring and processing neighbor information in machine learning. The 3D CNNs extract features that are subdivided into several levels in multiple dimensions, so they are particularly useful for identifying and predicting the cause of demand for items with large demand fluctuations or intermittent demand.

Second, previous studies have proposed image conversion and CNN architectures fixed to specific data and domains. However, in this study, we propose a method to train the architecture itself according to the dataset of the domain using the NAS. This is meaningful because Tab2vox is a general-purpose model that can be applied not only to the forecasting demand of items but also to various applications using tabular data. Furthermore, it can be extended to various tasks, such as classification and regression.

Third, since human intervention is minimized using the concept of NAS, which is an AutoML model, it can be a much more powerful model in real-world industrial settings.

*6.2. Limitations and Further Research*

Further studies can be conducted to overcome the limitations of this study:

1. It is only recently that researchers have begun to gather big data. We wanted to verify the Tab2vox framework in datasets from various industries, but it was difficult to acquire benchmark datasets segmented to a multi-level format. Additional verification by datasets from various domains is required.
2. After DARTS was first introduced, many NAS models with differentiable concepts appeared. With a more recent and advanced NAS model, the Tab2vox input embedding method could be further advanced.
3. Unlike previous studies using dimensionality reduction methodology, Tab2vox maps the features of each level to each dimension of the voxel image while preserving the dimension of the feature set. By applying the explainable artificial intelligence (XAI) technique, it is possible to add explainability to the model by performing actions such as displaying a heat map showing features on which the model focuses, and generating demand. Intuitive interpretation is possible because a single voxel itself has an explicit meaning.
4. Because of the intermittent demand characteristics, we had no choice but to use yearly data. In addition, despite using all datasets after 2010, when the Defense Logistics Integrated Information System (DELIIS) was developed, the time series period was not sufficient. Therefore, the target year was set to 1 year for analysis. The effect on mid- to long-term forecasting can also be reviewed with industrial data, such as production line data and electricity demand data, where demand occurs continuously rather than intermittently and the time series is sufficient.
5. In this work, the average effective image-embedding architecture and CNN architecture were searched for entire items by applying the Tab2vox framework to entire items. However, if the Tab2vox framework is applied to each item group, it will be possible to find a network that fits the characteristics of the item group. This is possible when sufficient samples for each item group are secured.

Key Points

- Tab2vox is a NAS model that embeds tabular data into optimal voxel images and finds the optimal CNN architecture for the images.
- We demonstrated better performance, compared to previous time series and machine learning techniques, using tabular data and the latest tabular to image conversion studies.
- Tab2vox is particularly meaningful because it is a general-purpose model that provides a framework that can be applied to demand forecasting in all sectors, rather than proposing a fixed architecture verified only in a specific dataset and domain.

**Supplementary Materials:** The following supporting information can be downloaded at: https://www.mdpi.com/article/10.3390/su141811745/s1, Figure S1: Architecture of the tab2vox demand forecasting framework; Figure S2: Result of item group analysis considering unit price of item and annual demand and its criteria.

**Author Contributions:** Conceptualization, E.L. and H.L.; methodology, E.L.; validation, E.L. and M.N.; formal analysis, E.L. and M.N.; investigation, E.L. and M.N.; resources, H.L.; data curation, E.L.; writing—original draft preparation, E.L.; writing—review and editing, E.L. and H.L.; visualization, E.L.; supervision, H.L. All authors have read and agreed to the published version of the manuscript.

**Funding:** This research received no external funding.

**Institutional Review Board Statement:** Not applicable.

**Informed Consent Statement:** Not applicable.

**Data Availability Statement:** The implementation of Tab2vox is available at https://github.com/kellyeunalee/tab2vox.

**Acknowledgments:** The authors would like to thank the anonymous reviewers for their valuable comments that helped revise the original version of this paper.

**Conflicts of Interest:** The authors declare no conflict of interest.